\documentclass{article}

\usepackage[preprint, nonatbib]{neurips_2022}

\usepackage[utf8]{inputenc} 
\usepackage[T1]{fontenc}    
\usepackage{hyperref}       
\usepackage{url}            
\usepackage{booktabs}       
\usepackage{amsfonts}       
\usepackage{nicefrac}       
\usepackage{microtype}      
\usepackage{xcolor}         
\usepackage{multirow}
\usepackage{graphicx}

\date{}

\usepackage{enumitem}

\usepackage{url}

\usepackage{hyperref} 

\begin{document}

\onecolumn 

\begin{description}[labelindent=-3cm,leftmargin=1cm,style=multiline]

\item[\textbf{Citation}]{C. Zhou, M. Prabhushankar, and G. AlRegib, "On the Ramifications of Human Label Uncertainty," in \textit{NeurIPS 2022 Workshop on Human in the Loop Learning}, New Orleans, LA, Dec. 2 2022.}


\item[\textbf{Review}]{Date of submission: Oct 7, 2022\\Date of acceptance: Oct 27, 2022}

\item[\textbf{Codes}]{\url{https://github.com/olivesgatech/Ramifications-HLU}}



\item[\textbf{Copyright}]{The authors grant NeurIPS a non-exclusive, perpetual, royalty-free, fully-paid, fully-assignable license to copy, distribute and publicly display all or parts of the paper. Personal use of this material is permitted. Permission from the authors must be obtained for all other uses, in any current or future media, including reprinting/republishing this material for advertising or promotional purposes, for resale or redistribution to servers or lists.}

\item[\textbf{Keywords}]{Human label uncertainty, Uncertainty quantification, Label dilution, Natural scene statistics} 

\item[\textbf{Contact}]{\href{mailto:chen.zhou@gatech.edu}{chen.zhou@gatech.edu}  OR 
 \href{mailto:mohit.p@gatech.edu}{mohit.p@gatech.edu} OR \href{mailto:alregib@gatech.edu}{alregib@gatech.edu}\\ \url{https://ghassanalregib.info/} \\ }
\end{description}

\thispagestyle{empty}
\newpage
\clearpage
\setcounter{page}{1}



\title{On the Ramifications of Human Label Uncertainty}

%

\author{%
  Chen Zhou \\
  Electrical and Computer Engineering\\
  Georgia Institute of Technology\\
  Atlanta, GA 30332-0250 \\
  \texttt{chen.zhou@gatech.edu} \\
  \And
  Mohit Prabhushankar \\
  Electrical and Computer Engineering \\
  Georgia Institute of Technology\\
  Atlanta, GA 30332-0250 \\
  \texttt{mohit.p@gatech.edu} \\
  \AND
  Ghassan AlRegib \\
  Electrical and Computer Engineering \\
  Georgia Institute of Technology \\
  Atlanta, GA 30332-0250 \\
  \texttt{alregib@gatech.edu} \\
}


\maketitle

\begin{abstract}
Humans exhibit disagreement during data labeling. We term this disagreement as \emph{human label uncertainty}. In this work, we study the ramifications of human label uncertainty (HLU). Our evaluation of existing uncertainty estimation algorithms, with the presence of HLU, indicates the limitations of existing uncertainty metrics and algorithms themselves in response to HLU. Meanwhile, we observe undue effects in predictive uncertainty and generalizability. To mitigate the undue effects, we introduce a novel natural scene statistics (NSS) based label dilution training scheme without requiring massive human labels. Specifically, we first select a subset of samples with low perceptual quality ranked by statistical regularities of images. We then assign separate labels to each sample in this subset to obtain a training set with diluted labels. Our experiments and analysis demonstrate that training with NSS-based label dilution alleviates the undue effects caused by HLU.  


\end{abstract}

\section{Introduction}
\label{introduction}
Raw data usually comes with separate labels collected from multiple imperfect human annotators. These separate labels  lead to a lack of consensus when they disagree with each other. The lack of consensus between human annotations exists in all data labeling processes and can be critical in various applications, e.g., image classification \cite{wei2021learning}, medical diagnosis \cite{ju2022improving, prabhushankar2022olives}, and seismic interpretation \cite{wang2019residual, benkert2022example}. We term this disagreement between human annotators as \emph{human label uncertainty}. 

The level of human label uncertainty (HLU) for a dataset is determined by the ratio of samples that are associated with multiple different human annotations. This ratio can be considered the label noise rate of the worst-case of aggregated human noisy labels, i.e., using one of any incorrect annotations as ground truth. Hence, we can study the effects of HLU in the worst-case human label noise setting.
Following the uncertainty evaluation protocols in prior works \cite{gal2016dropout, hendrycks2019augmix, zhou2021amortized, prabhushankar2022introspective}, we assess predictive uncertainty and generalizability using three uncertainty quantification algorithms under two types of distribution shift, including rotation and corruption. Typically, compared to a vanilla network, one would expect higher log-likelihood (LL) and lower Brier Scores (BS) from Monte Carlo (MC) Dropout \cite{gal2016dropout} and deterministic uncertainty quantification (DUQ) \cite{van2020uncertainty}.

To verify the ramifications of HLU, we compare generalizability and uncertainty between two training schemes, a) training with the worst-case human label noise using CIFAR-10-N \cite{wei2021learning} that reflects HLU, and b) training with ideal clean labels. We use the original CIFAR-10 training images for both training schemes. We use the aforementioned evaluation and summarize the test results on CIFAR-10-Rotations and CIFAR-10-C \cite{hendrycks2019benchmarking} in the first two rows (Human Noise and Clean) of both the top and bottom parts in Table \ref{Tab:ramifications_hlu}. The detailed experimental setup is elaborated in Section~\ref{experiments}.
We observe inconsistent trends of the discrepancy in certain uncertainty metrics, i.e., LL and BS, between the two training settings. This indicates that LL and BS do not properly capture HLU. Besides the limitations of existing metrics, MC Dropout and DUQ are not able to tackle HLU since they do not enhance the performance of uncertainty measures. Furthermore, we observe consistent trends of the degradation in predictive uncertainty and generalizability, which are measured by entropy and accuracy, respectively.  
Given this evidence of the undue effects, we further justify that such effects are caused by humans. In other words, the ramifications should not be replicable by introducing synthetic noisy labels. We introduce synthetic random noisy labels that are dependent on data instances with the same worst human label noise rate, since real-world HLU follows an instance-dependent pattern \cite{wei2021learning}. The discrepancy between the third row (Synthetic Noise) and the second row (Human Noise) in both parts of Table \ref{Tab:ramifications_hlu} verifies that the effects of HLU are caused by humans. To these ends, we motivate the objective to tackle the ramifications of HLU.

\begin{table}[t]
\centering
\caption{Evaluation results on CIFAR-10-Rotations (top part) and CIFAR-10-C (bottom part). The three uncertainty quantification algorithms are measured against different training schemes. The ramifications of human label uncertainty are detailed in Section~\ref{introduction}, and the results of different label dilution strategies are detailed in  Section~\ref{experiments}. Overall, our proposed NSS-based label dilution mitigates the ramifications of human label uncertainty.
 LL: Log-likelihood, BS: Brier Scores. All models are trained with the original CIFAR-10 training images but with different labels.}
\begin{tiny}
\begin{tabular}{ccccc|cccc|cccc}
\hline
\multirow{4}{*}{Training Schemes} & \multicolumn{12}{c}{CIFAR-10-Rotations}                               \\ \cline{2-13} 
                                 & \multicolumn{4}{c}{Vanilla} & \multicolumn{4}{c}{MC Dropout \cite{gal2016dropout} } & \multicolumn{4}{c}{DUQ \cite{van2020uncertainty} } \\
                                 & \begin{tabular}{@{}c@{}}LL \\ $(\uparrow)$\end{tabular}      & \begin{tabular}{@{}c@{}}BS \\ $(\downarrow)$\end{tabular}      & \begin{tabular}{@{}c@{}}Entropy \\ $(\downarrow)$\end{tabular}      & \begin{tabular}{@{}c@{}}ACC \\ $(\uparrow)$\end{tabular}      & \begin{tabular}{@{}c@{}}LL \\ $(\uparrow)$\end{tabular}      & \begin{tabular}{@{}c@{}}BS \\ $(\downarrow)$\end{tabular}      & \begin{tabular}{@{}c@{}}Entropy \\ $(\downarrow)$\end{tabular}      & \begin{tabular}{@{}c@{}}ACC \\ $(\uparrow)$\end{tabular}        & \begin{tabular}{@{}c@{}}LL \\ $(\uparrow)$\end{tabular}      & \begin{tabular}{@{}c@{}}BS \\ $(\downarrow)$\end{tabular}      & \begin{tabular}{@{}c@{}}Entropy \\ $(\downarrow)$\end{tabular}      & \begin{tabular}{@{}c@{}}ACC \\ $(\uparrow)$\end{tabular}    \\ \hline 
 &  \multicolumn{12}{c}{\textbf{Ramifications of Human Label Uncertainty}} \\   \hline
Clean        & -2.811        & 0.905     &  0.675       & 0.446     & -2.746         & 0.901       & 0.696    & 0.446       & -2.416      & 0.856   &   1.071     & 0.427   \\  
Human Noise \cite{wei2021learning}            & -1.916        & 0.784     &      2.398       & 0.355      & -1.931         & 0.789      &  2.413      & 0.350      & -2.648      & 0.997    & 1.697    & 0.282    \\  
Synthetic Noise     &   -2.591     &  0.940    & 1.572        & 0.288      &   -2.640     &   0.954       & 1.585         & 0.276      &  -3.030      &   1.191      & 1.208      & 0.224      \\ \hline 
&  \multicolumn{12}{c}{\textbf{Label Dilution Strategies}} \\  \hline

Human-based             & -2.013        &       0.773       & 2.069                & 0.392              &  -2.010               &   0.773          & 2.058                        & 0.394              &    -2.028         &    0.784        & 2.068                       & 0.379        \\

Random-based            & -2.489        &       0.831       & 1.964                & 0.361              &  -2.566               &   0.842          & 1.933                       & 0.355              &    -2.530         &    0.883        & 1.722                       & 0.326       \\   


Density-based \cite{guo2018curriculumnet}        &  -3.657              &  0.985        & 0.928                & 0.344                  &   -3.556         &   0.979        & 0.942                       & 0.347                    &     -2.656            &   0.895        & 1.369                       & 0.347       \\ 
\textbf{NSS-based (ours)}             &  -3.068           &      0.918      & {1.264}                & {0.362}                &   -3.006          &   0.917         & {1.284}                       & {0.357}             &   -2.457        &   0.872     & {1.569}                       & {0.350}       \\
\hline 

\hline
\multirow{4}{*}{Training Schemes} & \multicolumn{12}{c}{CIFAR-10-C}                               \\ \cline{2-13} 
                                 & \multicolumn{4}{c}{Vanilla} & \multicolumn{4}{c}{MC Dropout \cite{gal2016dropout} } & \multicolumn{4}{c}{DUQ \cite{van2020uncertainty} } \\
                                 & \begin{tabular}{@{}c@{}}LL \\ $(\uparrow)$\end{tabular}      & \begin{tabular}{@{}c@{}}BS \\ $(\downarrow)$\end{tabular}      & \begin{tabular}{@{}c@{}}Entropy \\ $(\downarrow)$\end{tabular}      & \begin{tabular}{@{}c@{}}ACC \\ $(\uparrow)$\end{tabular}      & \begin{tabular}{@{}c@{}}LL \\ $(\uparrow)$\end{tabular}      & \begin{tabular}{@{}c@{}}BS \\ $(\downarrow)$\end{tabular}      & \begin{tabular}{@{}c@{}}Entropy \\ $(\downarrow)$\end{tabular}      & \begin{tabular}{@{}c@{}}ACC \\ $(\uparrow)$\end{tabular}        & \begin{tabular}{@{}c@{}}LL \\ $(\uparrow)$\end{tabular}      & \begin{tabular}{@{}c@{}}BS \\ $(\downarrow)$\end{tabular}      & \begin{tabular}{@{}c@{}}Entropy \\ $(\downarrow)$\end{tabular}      & \begin{tabular}{@{}c@{}}ACC \\ $(\uparrow)$\end{tabular}    \\ \hline  
&  \multicolumn{12}{c}{\textbf{Ramifications of Human Label Uncertainty}} \\   \hline
Clean        & -1.115        & 0.400   & 0.384    & 0.750     & -1.121         & 0.407       & 0.396     & 0.746       & -1.161      & 0.451  &  0.829    & 0.686   \\  
Human Noise \cite{wei2021learning}            & -1.150        & 0.504      &  2.172   & 0.640      & -1.170         & 0.511     &  2.184     & 0.634      & -1.845       & 0.735   &   1.563      & 0.470       \\  
Synthetic Noise     &   -1.615     &  0.679     & 1.387        & 0.530        &   -1.639        &   0.690    & 1.399         & 0.514      &  -2.180    &  0.948     & 1.211      & 0.357      \\ \hline  
&  \multicolumn{12}{c}{\textbf{Label Dilution Strategies}} \\  \hline
Human-based             & -1.020        &       0.442        & 1.855               & 0.702              &  -1.002               &   0.435           & 1.848                       & 0.706              &    -1.104         &    0.472        & 1.863                        & 0.664       \\   

Random-based            & -1.381        &       0.563       & 1.765                 & {0.634}               &  -1.459               &   0.580          & 1.747                       & {0.618}              &    -1.378         &    0.580        & 1.583                       & {0.566}       \\  

Density-based \cite{guo2018curriculumnet}        &  -1.907              &  0.605        & 0.736                 & 0.582                   &   -1.830         &   0.592        & 0.743                       & 0.590                    &     -1.461             &   0.568         & 1.144                       & 0.575       \\ 
\textbf{NSS-based (ours)}             &  -1.583           &      0.594      & {1.222}                 & 0.580                &   -1.568          &   0.596         & {1.243}                       & 0.580              &   -1.392         &   0.591     & {1.496}                       & 0.540       \\
\hline 
\end{tabular}
\end{tiny}
\label{Tab:ramifications_hlu}
\end{table}

To address the undue effects caused by HLU, the authors in \cite{wei2022aggregate} show that training with separate human labels can be beneficial. However, collecting separate annotations from multiple independent human annotators can be challenging and expensive. In order to avoid collecting massive subjective human uncertain labels, we aim to generate separate labels in an objective manner for label dilution training. 
The cause of certain samples that exhibit HLU relates to the difficulty in perceiving image signals by humans. Perceptual quality assessment \cite{moorthy2011blind, temel2016unique, seijdel2020low, prabhushankar2017ms}, which can be objectively quantified by natural scene statistics (NSS), measures such difficulty. Hence, we propose an NSS-based label dilution training scheme to mitigate the undue effects caused by HLU. Specifically, we utilize statistical regularities of image signals to target which samples are likely to exhibit HLU. For each of these targeted samples, we then assign different labels to perform label dilution.
We demonstrate that the undue effects caused by HLU can be mitigated by our proposed NSS-based label dilution. In contrast to training with separate human labels, our approach does not require massive human annotations. We also compare against a  learning-based label dilution as an alternative. Our NSS-based label dilution, without training a perceptual learning model, achieves a comparable mitigation impact on the undue effects caused by HLU. In order to further verify the efficacy of utilizing NSS for label dilution, we conduct an ablation study that randomly targets training samples to assign separate labels. The results of the ablation study demonstrate the contributions of the NSS-based label dilution in alleviating the undue effects of HLU.  Contributions of our work include (\textit{i}) exposing the ramifications of HLU in terms of predictive uncertainty and generalizability,  ($ii$) identifying the limitations of certain existing uncertainty metrics and uncertainty estimation algorithms in response to HLU, and (\textit{iii}) introducing a novel NSS-based label dilution training scheme to tackle the ramifications of HLU without requiring massive human annotations.

\section{Methodology}
\label{methodology}
In this section, we present a novel NSS-based label dilution training scheme that tackles the ramifications of HLU. Specifically, the label dilution is applied to a subset of training samples that are targeted via statistical regularities of images. 

\begin{figure}[t]
\centering
\includegraphics[width=\columnwidth]{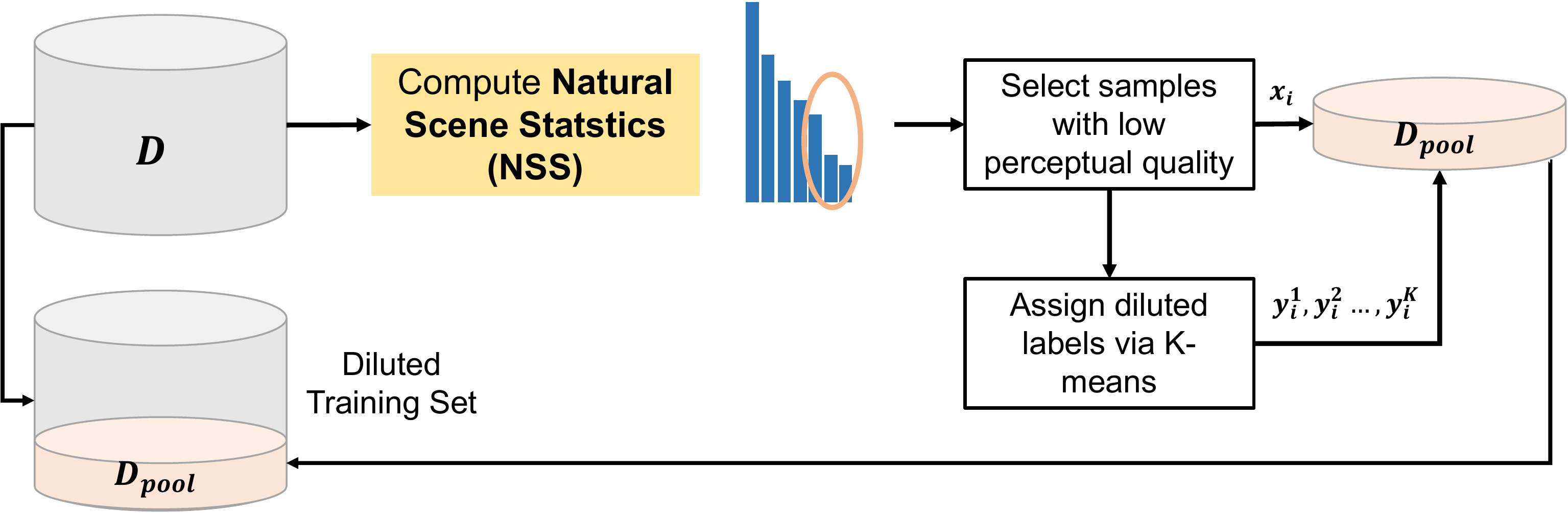}

\caption{Diagram of natural scene statistics (NSS) based label dilution. We sample a subset $D_{pool}$ according to the NSS scores to perform label dilution in an objective manner. Our NSS-based label dilution does not require massive human annotations.  \vspace{-.3cm}}
\label{fig: diagram_nss_dilution}
\end{figure}

The overall diagram of our proposed label dilution scheme is illustrated in Figure ~\ref{fig: diagram_nss_dilution}. Consider a network $f (\cdot)$, a training dataset $D$. We inject diluted labels into a subset pool $D_{pool}$ from $D$. Specifically, each sample $x_i\in D_{pool}$ is associated with a set of separate labels, i.e., $y_i^1, ..., y_i^k, k\in \{1,2,..., K\}$, where $K$ denotes the cardinality of the set of separate labels. The remaining samples are associated with clean labels. Note that for a sample $x_i \in D_{pool}$, the network $f (\cdot)$ is given its $k$ replications, each associated with one of the labels drawn from $y_i^1, ..., y_i^k$.  

One major question to address is how to determine $D_{pool}$. Annotations collected from multiple human annotators provide a solution. However, it is challenging and expensive to obtain massive human annotations. Hence, we propose to exploit natural scene statistics extracted from image samples to objectively determine $D_{pool}$.
Particularly, we employ BRISQUE \cite{mittal2012no}, an NSS-based algorithm, to obtain perceptual quality scores of all training images. Given an image sample, BRISQUE \cite{mittal2012no} exploits its spatial statistics of locally normalized luminance coefficients to quantify the perceptual quality. Samples that are in a relatively low-quality range are likely to exhibit relatively high HLU. We target this subset of training samples as $D_{pool} $ for label dilution. 
Once $D_{pool}$ is determined, the next question is how to assign multiple separate labels to each sample for label dilution. Moderately uncertain samples are likely to be associated with semantically relevant noisy labels, instead of random labels that are completely semantically irrelevant. Hence, we fit a $K$-means model on the training data and assign multiple labels to each sample $x_i \in D_{pool}$ based on its distance to the clustering centroids. We combine $D_{pool}$ with the remaining clean data to train a network $f (\cdot)$.

\section{Experiments}
\label{experiments}
We aim to address the following two questions: 1) Does NSS-based label dilution mitigate the ramifications of HLU? 2) How does NSS-based label dilution compare to other alternatives?
To answer these two questions, we conduct comparative experiments on evaluating a ResNet-18 \cite{he2016deep} that is trained with our proposed NSS-based label dilution scheme and two alternatives, including density-based and random-based dilution.

\textbf{Experimental setup} 
We experiment with three uncertainty quantification (UQ) techniques, including a vanilla ResNet18 (Vanilla), MC Dropout \cite{gal2016dropout}, and DUQ \cite{van2020uncertainty}. We include LL and BS for completeness, however, they are unreliable to reflect the ramifications of HLU. Hence,  we also evaluate other measures, e.g., entropy \cite{shannon1948mathematical}, a metric that does not rely on ground truth, to quantify the model's predictive uncertainty. For all UQ algorithms, we use the same base ResNet-18 \cite{he2016deep}. All models are evaluated on two outlier test sets, including a distortion test set CIFAR-10-C \cite{hendrycks2019benchmarking} and a rotated version of the CIFAR-10 test set (CIFAR-10-R). To create CIFAR-10-R, we rotate every image in the CIFAR-10 test set by $\gamma$, where $\gamma$ ranges from $0$ to $180$ degree with $12$ degree intervals.  
The training samples consist of the original CIFAR-10 \cite{krizhevsky2009learning} training images, some of which are associated with diluted labels in $D_{pool}$. To construct $D_{pool}$, we ensure that the ratio of diluted labels is the same as the worst-case human label noise rates in CIFAR-10-N \cite{wei2021learning}, which contains real-world human noisy annotations. 
Specifically, 
we fix the size of $D_{pool}$ to be $40\%$ size of the training set.
All samples in $D_{pool}$ have a maximum $K=3$ separate labels. For each sample in $D_{pool}$ with the top $10\%$ uncertainty, we dilute with at least two noisy labels. The remaining $30\%$ of samples are diluted with one noisy label. In this way, we construct $D_{pool}$ to have the same overall label disagreement configuration as that of the real-world human annotations in \cite{wei2021learning}.

\textbf{Results and analysis}
The results of our NSS-based label dilution strategy and two alternatives are summarized in the last four rows (Label Dilution Strategies) in both parts of Table \ref{Tab:ramifications_hlu}. Overall, the increase in entropy measures, caused by HLU in the second row (Human Noise), can be mitigated by our NSS-based label dilution training scheme as shown in the last row (NSS-based (ours)). In addition to mitigating the increase in predictive uncertainty, training with our NSS-based label dilution also enhances robustness as measured in classification accuracy (ACC). This observation provides the answer to the aforementioned first question, that our NSS-based label dilution mitigates the ramifications of HLU. We also include the results of human-based label dilution using real-world human annotations \cite{wei2021learning}. While human-based dilution achieves the highest generalizability, it requires massive human annotations.   
To answer the aforementioned second question, we alternate the sample selection procedure that determines $D_{pool}$, while following the same diluted label assignment procedure as in our NSS-based dilution.
In addition to diluting labels via NSS, one can also dilute labels based on the complexity of  samples \cite{guo2018curriculumnet} when projected to a particular feature distribution. Samples drawn from the same annotated category with a relatively lower feature density possess relatively large deviations in their visual statistics and are likely to exhibit HLU. Thus, we conduct the feature density-based label dilution as an alternative. Compared to the models trained with our NSS-based label dilution, the models trained with density-based dilution show lower uncertainty measured by entropy as illustrated in the second last row. One potential reason is that the density-based dilution enforces a more balanced category distribution of $D_{pool}$ since it conducts feature clustering in each category separately. While the density-based dilution fits a clustering model on samples for every class, our NSS-based dilution does not require fitting any learning-based models.
To develop an understanding of the efficacy of utilizing NSS to select samples that exhibit HLU, we conduct an ablation study that randomly selects the same amount of samples as $D_{pool}$.  All models trained with random-based label dilution show the largest uncertainty among the three objective label dilution strategies. While random-based label dilution can alleviate the ramifications of HLU in terms of classification accuracy, it can impair the models' uncertainty. This confirms the efficacy of utilizing NSS for label dilution to enhance the model's uncertainty.

\section{Discussion}
\label{discussion}
\textbf{Limitations and future work} This work illustrates the benefits of utilizing NSS for label dilution to mitigate the undue effects caused by HLU. However, the NSS-based label dilution does not explicitly consider the ambiguity in category semantics. In contrast, the learning-based label dilution exploits category semantics. Thus, a more proper label dilution is likely to utilize both NSS and feature semantics. Additionally, identifying more reliable metrics to capture HLU is required in the future. Finally, we acknowledge that, in certain critical applications, e.g., medical diagnosis and evaluation for surgery, one would favor agreement between humans on an evaluation instead of diluting labels.

\textbf{Conclusion} In this work, we investigate the ramifications of human label uncertainty (HLU). We observe that certain existing uncertainty metrics and algorithms are insufficient in response to HLU. Meanwhile, we expose the undue effects on the model's predictive uncertainty and generalizability caused by {HLU}. We then demonstrate that the undue effects can be mitigated by our proposed NSS-based label dilution training scheme, which does not require massive human labels. We conduct an ablation study to verify the efficacy of utilizing NSS for label dilution. 

\bibliographystyle{plain}
\bibliography{neurips_2022}

\newpage

\section*{Checklist}


\begin{enumerate}

\item For all authors...
\begin{enumerate}
  \item Do the main claims made in the abstract and introduction accurately reflect the paper's contributions and scope?
    \answerYes{}
  \item Did you describe the limitations of your work?
    \answerYes{See Section~\ref{discussion}.}
  \item Did you discuss any potential negative societal impacts of your work?
    \answerYes{See Section~\ref{discussion}.}
  \item Have you read the ethics review guidelines and ensured that your paper conforms to them?
    \answerYes{}
\end{enumerate}

\item If you are including theoretical results...
\begin{enumerate}
  \item Did you state the full set of assumptions of all theoretical results?
    \answerNA{}
        \item Did you include complete proofs of all theoretical results?
    \answerNA{}
\end{enumerate}

\item If you ran experiments...
\begin{enumerate}
  \item Did you include the code, data, and instructions needed to reproduce the main experimental results (either in the supplemental material or as a URL)?
    \answerNA{We will release the code upon paper acceptance.}
  \item Did you specify all the training details (e.g., data splits, hyperparameters, how they were chosen)?
    \answerNA{}
        \item Did you report error bars (e.g., with respect to the random seed after running experiments multiple times)?
    \answerNA{}
        \item Did you include the total amount of compute and the type of resources used (e.g., type of GPUs, internal cluster, or cloud provider)?
    \answerNA{}
\end{enumerate}

\item If you are using existing assets (e.g., code, data, models) or curating/releasing new assets...
\begin{enumerate}
  \item If your work uses existing assets, did you cite the creators?
    \answerYes{See References.}
  \item Did you mention the license of the assets?
    \answerNA{}
  \item Did you include any new assets either in the supplemental material or as a URL?
    \answerNA{}
  \item Did you discuss whether and how consent was obtained from people whose data you're using/curating?
    \answerNA{}
  \item Did you discuss whether the data you are using/curating contains personally identifiable information or offensive content?
    \answerNA{The datasets are open source.}
\end{enumerate}

\item If you used crowdsourcing or conducted research with human subjects...
\begin{enumerate}
  \item Did you include the full text of instructions given to participants and screenshots, if applicable?
    \answerNA{}
  \item Did you describe any potential participant risks, with links to Institutional Review Board (IRB) approvals, if applicable?
    \answerNA{}
  \item Did you include the estimated hourly wage paid to participants and the total amount spent on participant compensation?
    \answerNA{}
\end{enumerate}

\end{enumerate}


\appendix



\end{document}